\pdfoutput=1
\documentclass[11pt]{article}
\usepackage[table,xcdraw,dvipsnames]{xcolor}
\usepackage{enumerate}
\usepackage{times}
\usepackage{latexsym}
\usepackage[T1]{fontenc}
\usepackage[utf8]{inputenc}
\definecolor{darkorange}{rgb}{1, 0.549, 0}
\usepackage{microtype}
\usepackage{enumitem} 
\usepackage{bbding}
\usepackage{CJKutf8}
\usepackage{graphicx}
\usepackage{array}
\usepackage{arydshln}
\usepackage{multirow}

\usepackage{bm}
\usepackage{makecell}
\usepackage{wrapfig}
\usepackage{lipsum}  
\usepackage{etoolbox}
\usepackage{comment}
\usepackage[]{EMNLP2022}
\usepackage{booktabs}
  
\usepackage{csquotes}
\usepackage{bm}
\usepackage{amsmath,amssymb, amsfonts,mathtools}
\usepackage{soul}
\usepackage{adjustbox}
\usepackage[]{todonotes}
\usepackage{color, colortbl}
	
\definecolor{Gray}{gray}{0.9}
\definecolor{LightCyan}{rgb}{0.88,1,1}
\definecolor{LightBlue}{rgb}{236,244,255}
\usepackage{colortbl}
\usepackage{pgfplots}
\usepackage{hyperref}
\hypersetup{colorlinks,allcolors=black}
\AtBeginEnvironment{appendices}{\crefalias{section}{appendix}}
\BeforeBeginEnvironment{appendices}{\clearpage}

\usepackage{cleveref}
\crefname{section}{\S}{\S\S}
\crefname{table}{Tab.}{}
\crefname{figure}{Fig.}{Figs.}
\crefname{algorithm}{Algorithm}{}
\crefname{equation}{Eq.}{Eqs.}
\crefname{line}{Line}{}
\crefname{appendix}{App.}{}
\crefformat{section}{\S#2#1#3}
\crefname{thm}{Theorem}{}
\crefname{cor}{Corollary}{}
\crefname{prop}{Proposition}{}
\crefname{def}{Definition}{}

\usepackage{scalerel}
\usepackage[ruled,vlined]{algorithm2e}
\usepackage{linguex}
\usepackage{relsize}
\usepackage{array}
\usepackage{makecell}

\usepackage[font=small,labelfont=bf]{caption}
\usepackage{subcaption}
\usepackage{cancel}
\usepackage{inconsolata}

\usepackage{appendix}
\BeforeBeginEnvironment{appendices}{\clearpage}

\usepackage{todonotes}

\definecolor{babyblue}{rgb}{0.54, 0.81, 0.94}

\usepackage{stmaryrd}

\newcommand{\Verb}[1]{\textcolor{teal}{\textbf{\textit{#1}}}}

\newcommand{\redEnt}[1]{\textcolor{red}{\textbf{#1}}}
\newcommand{\bleuEnt}[1]{\textcolor{blue}{\textbf{#1}}}
\newcommand{\pinkEnt}[1]{\textcolor{magenta}{\textbf{#1}}}
\newcommand{\cyanEnt}[1]{\textcolor{cyan}{\textbf{#1}}}
\newcommand{\Omit}[1]{$\llbracket$\textbf{#1}$\rrbracket$}

\newcommand{\redEntOmit}[1]{\textcolor{red}{\Omit{#1}}}
\newcommand{\bleuEntOmit}[1]{\textcolor{blue}{\Omit{#1}}}
\newcommand{\pinkEntOmit}[1]{\textcolor{magenta}{\Omit{#1}}}
\newcommand{\cyanEntOmit}[1]{\textcolor{cyan}{\Omit{#1}}}
\newcommand{\grayEnt}[1]{\textcolor{gray}{\textbf{#1}}}
\newcommand{\orangeEnt}[1]{\textcolor{darkorange}{\textbf{\underline{#1}}}}

\newcommand{\NoError}{\textsc{no error}}
\newcommand{\Document}{\textsc{document}}
\newcommand{\Sentence}{\textsc{sentence}}

\newcommand{\Entity}{\textsc{entity}}
\newcommand{\Tense}{\textsc{tense}}
\newcommand{\Ellipsis}{\textsc{ellipsis}}
\newcommand{\Pronoun}{\textsc{pronoun}}
\newcommand{\ZeroPronoun}{\textsc{zeroPro}}

\newcommand{\Ambiguity}{\textsc{ambiguity}}

\newcommand{\Masculine}{\textsc{masculine}}
\newcommand{\Feminine}{\textsc{feminine}}
\newcommand{\Neuter}{\textsc{neuter}}
\newcommand{\Epicene}{\textsc{epicene}}

\newcommand{\BWB}{$\mathcal{BWB}$}

\newcommand{\MT}{\textsc{mt}}

\newcommand{\SMT}{\textsc{smt}}
\newcommand{\OMTa}{\textsc{ggl}}
\newcommand{\OMTb}{\textsc{bd}}
\newcommand{\OMTc}{\textsc{bing}}
\newcommand{\MTS}{\textsc{mt-s}}
\newcommand{\MTD}{\textsc{mt-d}}
\newcommand{\HT}{\textsc{ht}}

\newcommand{\PE}{\textsc{pe}}
\newcommand{\bsc}[1]{\textbf{\textsc{#1}}}

\newcommand{\BLEU}{\textsc{bleu}}
\newcommand{\BlonD}{\textsc{blonde}}

\newcommand{\BERTScore}{\textsc{bert}}
\newcommand{\METEOR}{\textsc{meteor}}
\newcommand{\TER}{\textsc{ter}}

\newcommand{\fluency}{\textsc{fluency}}
\newcommand{\adequacy}{\textsc{adequacy}}
\newcommand{\testset}[1]{\textsc{part#1}}
\newcommand{\rater}[1]{\textsc{rater#1}}

\newcommand{\myFontSize}{\scriptsize}
\def\stripzero#1{\expandafter\stripzerohelp#1}
\def\stripzerohelp#1{\ifx 0#1\expandafter\stripzerohelp\else#1\fi}
\newcommand{\ZH}[1]{{\myFontSize \begin{CJK*}{UTF8}{gbsn} #1 \end{CJK*}}}

\newcommand{\ethz}{\textsuperscript{$\zeta$}}
\newcommand{\msra}{\textsuperscript{$\gamma$}}

\title{A Bilingual Parallel Corpus with Discourse Annotations}
\author{
Yuchen Eleanor Jiang{\ethz}~\;~Tianyu Liu{\ethz}~\;~Shuming Ma{\msra}\\
\textbf{Dongdong Zhang{\msra}~\;~Mrinmaya Sachan{\ethz}~\;~Ryan Cotterell{\ethz}}\\
  $^{\ethz}$ETH Z\"{u}rich~\;~$^{\msra}$Microsoft Research Asia\\
    \texttt{\{\href{mailto:yuchen.jiang@inf.ethz.ch}{yuchen.jiang},\href{mailto:tianyu.liu@inf.ethz.ch}{tianyu.liu},\href{mailto:ryan.cotterell@inf.ethz.ch}{ryan.cotterell},\href{mailto:mrinmaya.sachan@inf.ethz.ch}{mrinmaya.sachan}\}@inf.ethz.ch }\\
\texttt{\{\href{mailto:shuming.ma@microsoft.com}{shuming.ma},\href{mailto:dongdong.zhang@microsoft.com}{dongdong.zhang}\}@microsoft.com}
}
\date{}
\begin{document}
\maketitle

\begin{abstract}
Machine translation (MT) has almost achieved human parity at sentence-level translation.
In response, the MT community has, in part, shifted its focus to document level translation.
However, the development of document-level MT systems is hampered by the lack of parallel document corpora.
This paper describes \BWB{}, a large parallel corpus first introduced in \citet{jiang-etal-2022-blonde}, along with an annotated test set.
The \BWB{} corpus consists of Chinese novels translated by experts into English, and the annotated test set is designed to probe the ability of machine translation systems to model various discourse phenomena.
Our resource is freely available, and we hope that it will serve as a guide and inspiration for more work in the area of document-level machine translation.

\noindent {\includegraphics[width=1.25em,height=1.25em]{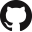}\hspace{.75em}\parbox{\dimexpr\linewidth-2\fboxsep-2\fboxrule}{\url{https://github.com/EleanorJiang/BlonDe/tree/main/BWB}}}
\end{abstract}

\section{Introduction}
Machine translation (MT) has made significant progress in the past few decades.
Neural machine translation (NMT) models, which are able to leverage abundant quantities of parallel training data, have been one of the main contributors to this progress~\citep[][\textit{inter alia}]{luong-etal-2015-effective,transformer,ctx}.
Unfortunately, the majority of available parallel corpora contain sentence level translations. 
As a result, models trained on these corpora translate text quite well at the sentence level, but perform poorly when the entire document translation is seen in context \citep{voita-etal-2019-good, werlen2021discourse}.
Particularly, sentence level translation models 
 tend to omit relevant contextual information, resulting in a lack of coherence 
in the produced translation. 
For example, in \Cref{fig:intro_example}, the sentence-level MT system fails to capture discourse dependencies across sentences and the same concepts are not consistently referred to with the same translations (i.e. \orangeEnt{Weibo} vs \orangeEnt{micro-blog}, \redEnt{Qiao Lian} vs \redEnt{Joe} vs \redEnt{Joe love}).\footnote{This discourse phenomenon is referred as entity consistency. There are other discourse dependencies that MT fails to capture, such as tense cohesion, ellipsis and coreference. We have left explanations of discourse phenomena to \Cref{subsec:error}.}

\begin{figure}[t]
    \centering
    \vspace{-10pt}
    \includegraphics[width=0.4\textwidth,trim={1cm 7cm 1cm 1.6cm} ,clip]{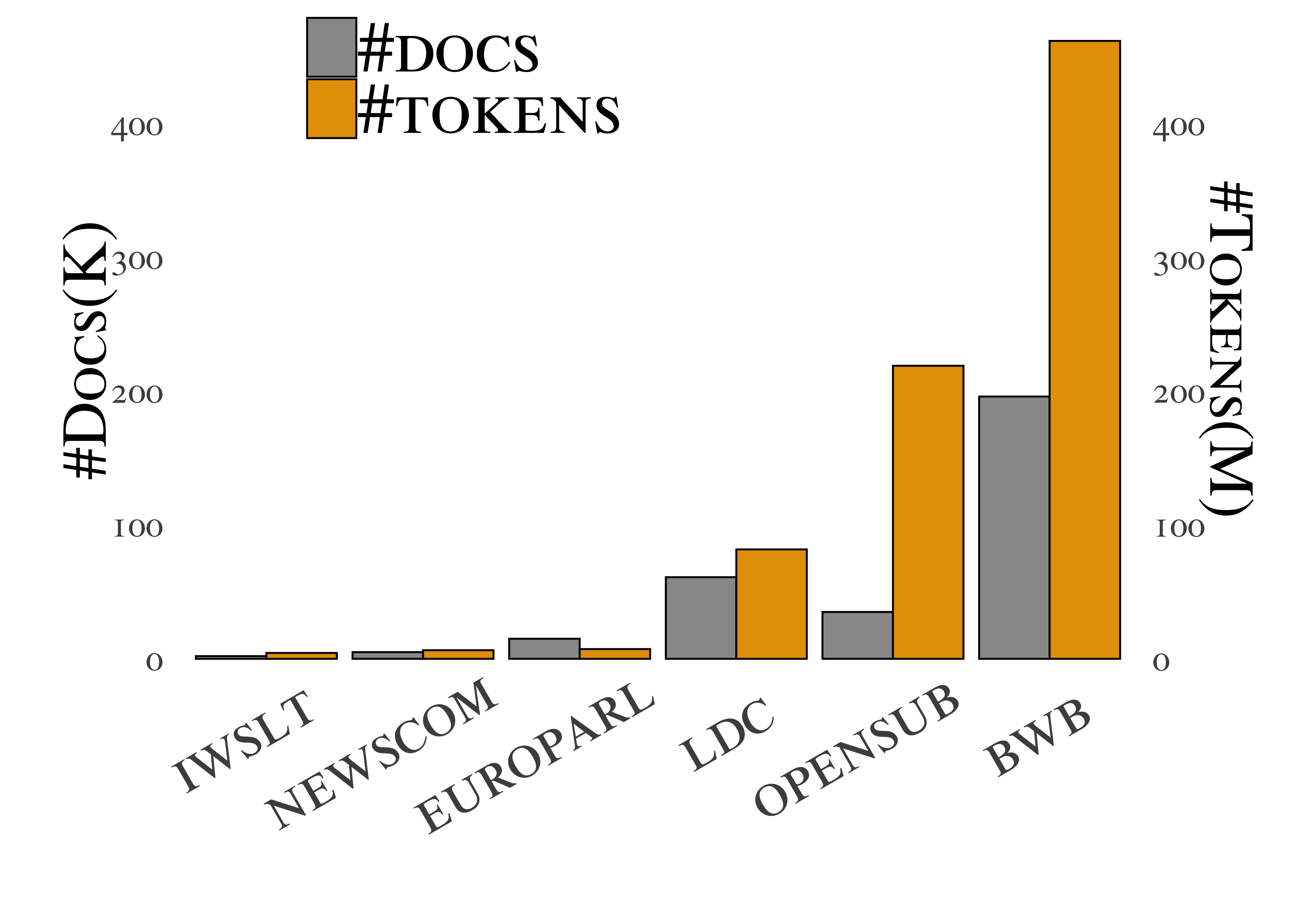}
    \vspace{-6pt}
    \caption{Comparing sizes of various document-level parallel corpora. \BWB{} is the to-date largest parallel corpus. \looseness=-1}
    \vspace{-10pt}
    \label{fig:motivation}
\end{figure}

\begin{figure*}[t]
\begin{adjustbox}{width=\textwidth}
\centering \scriptsize
\begin{tabular}{p{5pt}p{150pt}p{155pt}p{150pt}}
\toprule[2pt]
& SOURCE & REFERENCE & MT \\
\midrule[1pt]
1) & \ZH{	\redEnt{乔恋}攥紧了拳头，垂下了头。	}&	\redEnt{Qiao Lian} clenched \redEnt{her} fists and lowered \redEnt{her} head.	&	\redEnt{Joe} clenched \redEnt{his} fist and bowed \redEnt{his} head.	\\
2) & \ZH{	其实\bleuEnt{他}说得对。	}&	Actually, \bleuEnt{he} \Verb{was} right.	&	In fact, \bleuEnt{he}\Verb{'s} right.	\\
3) & \ZH{	\redEntOmit{}自己就是一个蠢货，竟然会[...]。	}&	\redEntOmit{She} \Verb{was} indeed an idiot, as only an idiot \Verb{would} [...]	&	\redEntOmit{I} \Verb{am} a fool, even \Verb{will} [...]	\\
5) & \ZH{	\redEnt{她}点进去，发现是\orangeEnt{凉粉群}，所有人都在@\redEnt{她}[...]	}&	\redEnt{She} logged into \redEntOmit{her} account and saw that a large number of fans in the \orangeEnt{Liang fan group} had tagged \redEnt{her}.[...]	&	\redEnt{She} nodded in and found it was a \orangeEnt{cold powder group}, and everyone was on \redEnt{her}.[...]	\\
7) & \ZH{	【\grayEnt{川流不息}：\redEnt{乔恋}，快看\orangeEnt{微博}头条！ \orangeEnt{微博}头条！】	}&	[\grayEnt{Chuan Forever}: \redEnt{Qiao Lian}, look at the headlines on \orangeEnt{Weibo}, quickly!]	&	\grayEnt{Chuan-flowing}: \redEnt{Joe love}, quickly look at the \orangeEnt{micro-blogging} headlines! \orangeEnt{Weibo} headlines?	\\
8) & \ZH{	\redEnt{她}微微一愣，拿起手机，登陆\orangeEnt{微博}，在看到头条的时候，整个人一下子愣住了！	}&	\redEnt{She} froze momentarily, then picked up \redEntOmit{her} cell phone and logged into \orangeEnt{Weibo}. When  \redEntOmit{she} saw the headlines,  \redEntOmit{her entire body} immediately froze over again!	&	\redEnt{She} took a slight look, picked up the phone, landed on the \orangeEnt{micro-blog}, when  \redEntOmit{she} saw the headlines,  \redEntOmit{the whole person} suddenly choked! \\
\bottomrule[2pt]
\end{tabular}
\end{adjustbox}
\caption{Part of a chapter in \BWB. The same entities are marked with the same color. 
Pronoun omissions are marked with \Omit{}.
The mistranslated verbs are marked with \Verb{teal},
and the mistranslated named entities are \orangeEnt{underlined}. The full chapter is in \Cref{fig:example_book1_0}. MT is the output of a Transformer-based sentence-level machine translation system. \looseness=-1}
\label{fig:intro_example}
\vspace{-5pt}
\end{figure*}

Over the past few years, there have been efforts to tackle this problem by building context-aware NMT models~\citep[][\textit{inter alia}]{wang-etal-2017-exploiting, miculicich-2018-hat, maruf-haffari-2018-document, voita-etal-2019-context}. 
Although such approaches have achieved some improvements, they nonetheless suffer from a dearth of document level training data. 
Take the WMT news translation task as an example. The document level news commentary corpus~\cite{tiedemann-2012-OPUS} only contains 6.4M tokens while the available sentence-level training data has around 825M tokens.
To alleviate this problem, we collect a large document level parallel corpus that consists of 196K paragraphs from Chinese novels translated into English. 
As shown in \Cref{fig:motivation}, it is the largest document-level corpus to the best of our knowledge.
Additionally, an in-depth human analysis shows, it is very challenging for current NMT systems due to its rich discourse phenomena (see \Cref{fig:intro_example}).

To better evaluate context-aware MT models, we further annotate the test set with characteristic discourse-level phenomena, namely ambiguity and ellipsis.
The test set is designed to specifically measure models’ capacity to exploit such long range linguistic context. 
We then conduct systematic evaluations of several baseline models as well as human post-editing performance on the \BWB{} corpus and observe large gaps between NMT models and human performance.
We hope that this corpus will help us understand the deficiencies of existing systems and build better systems for document level machine translation.

\begin{table}[t]
\begin{adjustbox}{width=0.49\textwidth}
{
\centering
\begin{tabular}{l|l|rrr|rrr}
\toprule[2pt]
\multirow{2}{*}{Corpus}  & \multirow{2}{*}{Genre}  & \multicolumn{3}{c|}{Size}                                                                         & \multicolumn{3}{c}{Averaged Length}                    \\
                    &                           & \#word & \#sent & \#doc & \#w/s & \#s/d & \#w/d \\
\midrule[1pt]
	
IWSLT  & TED talk &  4.2M                         & 0.2M                            & 2K                              & 19.5                           & 100                            & 2,100                         \\
NewsCom & news    & 6.4M                         & 0.2M                            & 5K                              & 30.7                           & 40                             & 1,288                         \\
Europarl                & Parliament            & 7.3M                         & 0.2M                            & 15K                             & 35.1                           & 13                             & 485                           \\
LDC                     & News                      &  81.8M                        & 2.8M                            & 61K                             & 23.7                           & 46                             & 1,340                         \\
OpenSub &       Subtitle     &  16.9M                        & 2.2M                            & 3K                              & 5.6                            & 733                            & 5,647                         \\
\midrule[1pt]
\multirow{2}{*}{\BWB}     & novel (chapter)           &  \textbf{460.8M}                       & \textbf{9.6M}                            & \textbf{196K}                            & \textbf{48.1}                          & 49                             & 2,356                         \\
&    novel (book)              &  \textbf{460.8M}                       & \textbf{9.6M}                            & 384                             & \textbf{48.1}                           & \textbf{25.0K}                          & \textbf{1.2M}          \\
\bottomrule[2pt]
\end{tabular}
}
\end{adjustbox}
    \caption{Statistics of various document-level parallel corpora. For corpora that contain multiple language pairs (IWSLT and OpenSub), we report the statistics for ZH-EN. For corpora that do not contain ZH-EN parallel documents (NewsComand and Europarl), we report the statistics of their (largest) available language pairs (DE-EN and ET-EN). \textit{w}, \textit{s} and \textit{d} stand for word, sentence and document, respectively. The full list is in \Cref{tab:datasets_full}.\looseness=-1}
    \label{tab:datasets}
\vspace{-10pt}
\end{table}

\section{Dataset Creation} \label{sec:datasetCreation}
In this section, we describe three stages of the dataset creation process: collecting bilingual parallel documents, quality control and dataset split.
\subsection{Bilingual Document Collection} \label{subsec:quality}
\begin{figure}[t]
    \centering
    \vspace{-8pt}
    \includegraphics[width=0.3\textwidth]{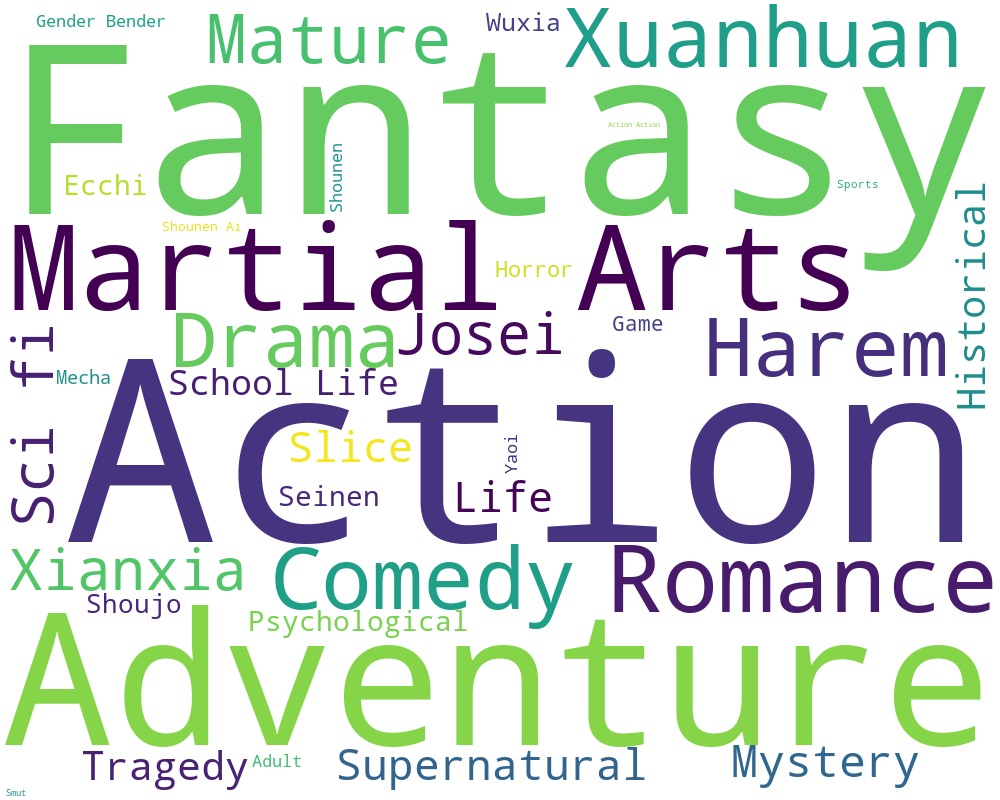}
    \vspace{-6pt}
    \caption{The genre distribution of novels in the \BWB{} corpus. Action, Fantasy and Adventure are the most common genres.\looseness=-1}
    \vspace{-10pt}
    \label{fig:genre_wordcloud}
\end{figure}

We first select 385 Chinese web novels across multiple genres, including action, fantasy, romance, comedy, science fictions, martial arts, etc. The genre distribution is shown in \Cref{fig:genre_wordcloud}. 
We then crawl their corresponding English translations from the Internet.\footnote{\url{https://readnovelfull.com}}
The English versions are translated by professional translators who are native speakers of English, and then corrected and aligned by professional editors at the chapter level.
The text is converted to UTF-8 and certain data cleansing (e.g. deduplication) is performed in the process. 
Chapters that contain poetry or couplets in classical Chinese are excluded as they are difficult to translate directly into English.
Further, we exclude chapters with less than 5 sentences and chapters where the sequence ratio is greater than 3.0.
The titles of each chapter are also removed, since most of them are neither translated properly nor at the document level.
The sentence alignment is automatically performed by Bleualign\footnote{\url{https://github.com/rsennrich/Bleualign}}~\cite{bleualign}. 
The final corpus has 384 books with 9,581,816 sentence pairs (a total of 461.8 million words).\footnote{We will release a crawling and cleansing script pointing to a past web arxiv that will enable others to reproduce our dataset faithfully.}

\begin{table}[t]
\centering
\begin{adjustbox}{width=0.49\textwidth}
\begin{tabular}{ccp{120pt}c}
\toprule[2pt]
Error Type & \# & Description & An\\
\midrule[1pt]
\Entity  & 43.3\% & error(s) due to the mistranslation of named entities. & \CheckmarkBold \\
\Tense  & 38.7\% & error(s) due to incorrect tense. & \\
\ZeroPronoun  & 17.3\% & error(s) caused by the omission of pronoun(s). & \CheckmarkBold \\
\Ambiguity  & 7.3\% & there are some ambiguous span(s) that is(are) correct in the stand-alone sentence but wrong in context. & \CheckmarkBold \\ 
\Ellipsis  & 4.0\% & error(s) caused by the omission of other span(s). & \CheckmarkBold \\
\Sentence & 51.3\% & sentence-level error(s). & \\
\NoError & 17.1\% & no errors. & \\
\bottomrule[2pt]
\end{tabular}
\end{adjustbox}
    \caption{The types of NMT errors and their description. \# represents the proportion of the error in the \BWB{} test set. \CheckmarkBold indicates ``with annotation''. \looseness=-1}
    \label{tab:errors}
    \vspace{-13pt}
\end{table}

\subsection{Quality Control} \label{subsec:quality}
We hired four bilingual graduate students to perform the quality control of the aforementioned process. %
These annotators were native Chinese speakers and proficient in English.
We randomly selected 163 chapters and asked the annotators to distinguish whether a document was well aligned at the sentence level by counting the number of misalignment. It is identified as a misalignment if, for example, line 39 in English corresponds to line 39 and line 40 in Chinese, but the tool made a mistake in combining the two sentences.
We observed an alignment accuracy rate of 93.1\%.

\subsection{Dataset Split} \label{subsec:split}
We construct the development set and the test set by randomly selecting 80 and 79 chapters from 6 novels, which contain 3,018 chapters in total.
To prevent any train-test leakage, these 6 novels are removed from the training set.
\Cref{tab:dataset_split} provides the detailed statistics of the \BWB{} dataset split.
In addition, we asked the same annotators who performed the quality control to manually correct misalignments in the development and test sets, and 7.3\% of the lines were corrected in total.

\section{Dataset Analysis and Annotation}\label{sec:analysis}
As part of this section, we analyze the types of translation errors that can occur in sentence-level NMT outputs, as well as annotate the \BWB{} test set.
We also provide analysis on coherence-related properties: numbers of named entities, numbers of pronouns in both English and Chinese, and the relationships of those factors.
The annotation was conducted by eight professional translators.

\subsection{Translation Errors} \label{subsec:error}
The annotators were asked to identify and categorize discourse-level translation errors made by a state-of-the-art commercial NMT system, i.e. errors that are only visible in context larger than individual sentences.
The annotators followed the following guideline for this error annotation:
\begin{enumerate}
    \vspace{-5pt}
    \item Identify cases that have translation errors: label examples as \NoError{} only if they meet both the criteria of adequacy and fluency as well as the global criterion of coherence.
    \vspace{-8pt}
    \item Identify whether the translation error is at the sentence level or document level (or both): \Sentence{} are examples that are already not adequate or fluent as stand-alone sentences. 
    \vspace{-8pt}
    \item Categorize the \Document{} examples in accordance with the discourse phenomena,  mark the corresponding spans 
    in the reference (English) that cause the MT output to be incorrect, and provide the correct versions.
    \vspace{-5pt}
\end{enumerate}
The types of errors are summarized in \Cref{tab:errors}.

\begin{table}[t]
\centering \small
\begin{adjustbox}{width=0.46\textwidth}
\begin{tabular}{ccccc}
\toprule[2pt]
Lang & \Masculine & \Feminine & \Neuter & \Epicene\\
\midrule[1pt]
EN & 1633 & 2521 & 608 & 391 \\
ZH & 654 & 967 & 14 & 118 \\
\bottomrule[2pt]
\end{tabular}
\end{adjustbox}
    \caption{The distributions of different types of pronouns in both English and Chinese in the \BWB{} test set. 
    \looseness=-1}
    \label{tab:pronoun}
    \vspace{-12pt}
\end{table}

\begin{table*}[t]
\vspace{-10pt}
\centering \small
\begin{adjustbox}{width=0.9\textwidth} 
\begin{tabular}{c|p{1cm}<{\centering}p{1cm}<{\centering}p{1cm}<{\centering}p{1cm}<{\centering}p{1cm}<{\centering}|cc|cc|cc|cc|cc}
\toprule[2pt]
 & \multicolumn{5}{c|}{Automatic Metrics} & \multicolumn{10}{c}{Discourse Phenomena} \\
& \BLEU & \METEOR & \TER &\BERTScore & \BlonD & \multicolumn{2}{c|}{\Ambiguity} & \multicolumn{2}{c|}{\Entity} & \multicolumn{2}{c|}{\Tense} & \multicolumn{2}{c|}{\Pronoun} & \multicolumn{2}{c}{\Ellipsis} \\
\midrule[1pt]
\SMT  & \cellcolor[HTML]{FDF8F8}6.86             & \cellcolor[HTML]{FEFEFE}18.73              & \cellcolor[HTML]{73A7D6}84.35           & \cellcolor[HTML]{FFFFFF}32.14                 & \cellcolor[HTML]{FEFEFE}15.52             & \cellcolor[HTML]{B4D0E9}28.92                 & \cellcolor[HTML]{C1D4E9}20.94 & \cellcolor[HTML]{D4E4F2}30.05              & \cellcolor[HTML]{E0EBF6}19.67 & \cellcolor[HTML]{FEFEFE}51.01             & \cellcolor[HTML]{FFFFFF}44.58 & \cellcolor[HTML]{FFFFFF}52.21               & \cellcolor[HTML]{FFFFFF}38.73 & \cellcolor[HTML]{FFFFFF}34.85                & \cellcolor[HTML]{FFFFFF}21.66 \\
\OMTb & \cellcolor[HTML]{CEDCEC}10.02            & \cellcolor[HTML]{3D85C6}26.57              & \cellcolor[HTML]{FEFEFE}70.72           & \cellcolor[HTML]{86B3DC}46.78                 & \cellcolor[HTML]{F9FBFD}16.17             & \cellcolor[HTML]{87B3DC}41.52                 & \cellcolor[HTML]{9DBEDF}31.47 & \cellcolor[HTML]{FEFEFE}21.28              & \cellcolor[HTML]{FEFEFE}12.04 & \cellcolor[HTML]{BAD3EB}58.36             & \cellcolor[HTML]{AFCDE8}54.19 & \cellcolor[HTML]{D7E6F4}56.81               & \cellcolor[HTML]{E3EDF7}43.65 & \cellcolor[HTML]{AFCDE8}52.53                & \cellcolor[HTML]{BED6EC}41.05 \\
\OMTc & \cellcolor[HTML]{ACC8E3}12.23            & \cellcolor[HTML]{8FB8DE}23.27              & \cellcolor[HTML]{3D85C6}89.53           & \cellcolor[HTML]{B2CFE9}41.45                 & \cellcolor[HTML]{C3D9ED}22.47             & \cellcolor[HTML]{FEFEFE}8.09                  & \cellcolor[HTML]{FDF8F8}3.33  & \cellcolor[HTML]{CDE0F0}31.37              & \cellcolor[HTML]{D4E4F2}22.62 & \cellcolor[HTML]{B7D2EA}58.66             & \cellcolor[HTML]{A9C9E6}54.93 & \cellcolor[HTML]{ADCCE7}61.54               & \cellcolor[HTML]{B0CEE8}52.29 & \cellcolor[HTML]{FFFFFF}34.88                & \cellcolor[HTML]{FFFFFF}21.74 \\
\OMTa & \cellcolor[HTML]{A3C2E1}12.81            & \cellcolor[HTML]{CCDFF0}20.80              & \cellcolor[HTML]{91BADF}81.39           & \cellcolor[HTML]{86B3DC}46.80                 & \cellcolor[HTML]{C0D7EC}22.86             & \cellcolor[HTML]{74A8D6}46.66                 & \cellcolor[HTML]{8DB5DB}36.05 & \cellcolor[HTML]{E0ECF6}27.46              & \cellcolor[HTML]{DFEBF5}20.03 & \cellcolor[HTML]{A1C4E3}61.01             & \cellcolor[HTML]{8BB6DD}58.58 & \cellcolor[HTML]{C3DAEE}59.04               & \cellcolor[HTML]{C6DBEF}48.63 & \cellcolor[HTML]{DAE8F5}43.04                & \cellcolor[HTML]{E1ECF6}30.76 \\
\MTS  & \cellcolor[HTML]{7EACD7}15.24            & \cellcolor[HTML]{A4C6E4}22.39              & \cellcolor[HTML]{AACAE6}78.93           & \cellcolor[HTML]{B1CEE8}41.58                 & \cellcolor[HTML]{9FC2E3}26.79             & \cellcolor[HTML]{4B8ECA}58.16                 & \cellcolor[HTML]{88B2DA}37.59 & \cellcolor[HTML]{FAFBFD}22.27              & \cellcolor[HTML]{DDE9F5}20.54 & \cellcolor[HTML]{4F91CC}69.77             & \cellcolor[HTML]{4E90CB}65.77 & \cellcolor[HTML]{8DB7DE}65.21               & \cellcolor[HTML]{74A8D6}62.64 & \cellcolor[HTML]{7BACD8}64.01                & \cellcolor[HTML]{8BB7DD}55.81 \\
\MTD  & \cellcolor[HTML]{5D98CF}17.45            & \cellcolor[HTML]{5F9AD0}25.21              & \cellcolor[HTML]{3E86C7}89.48           & \cellcolor[HTML]{97BDE1}44.78                 & \cellcolor[HTML]{76A9D7}31.53             & \cellcolor[HTML]{3D85C6}61.95                 & \cellcolor[HTML]{74A6D5}43.18 & \cellcolor[HTML]{DDE9F5}28.25              & \cellcolor[HTML]{C8DCEF}25.67 & \cellcolor[HTML]{4A8DCA}70.39             & \cellcolor[HTML]{3D85C6}67.80 & \cellcolor[HTML]{3D85C6}74.20               & \cellcolor[HTML]{3D85C6}71.96 & \cellcolor[HTML]{4F91CC}73.55                & \cellcolor[HTML]{5594CD}71.93 \\
\PE   & \cellcolor[HTML]{3D85C6}19.52            & \cellcolor[HTML]{A5C6E4}22.38              & \cellcolor[HTML]{AFCDE7}78.47           & \cellcolor[HTML]{3D85C6}55.50                 & \cellcolor[HTML]{3D85C6}38.18             & \cellcolor[HTML]{4288C8}60.65                 & \cellcolor[HTML]{3D85C6}59.09 & \cellcolor[HTML]{3D85C6}60.94              & \cellcolor[HTML]{3D85C6}60.41 & \cellcolor[HTML]{3D85C6}71.68             & \cellcolor[HTML]{4087C7}67.56 & \cellcolor[HTML]{5594CD}71.51               & \cellcolor[HTML]{7DAED9}61.04 & \cellcolor[HTML]{3D85C6}77.45                & \cellcolor[HTML]{3D85C6}78.75
\\
\bottomrule[2pt]
\end{tabular}
\end{adjustbox}
\vspace{-5pt}
\caption{Results of MT systems and human post-editing on the \BWB{} test set. For discourse phenomena, we report both F1 measure defined in \citet{blonde} and Exact-Match Accuracy defined in \citet{alam-terminology-2021}.  \looseness=-1}
\label{tab:scores}
\vspace{-10pt}
\end{table*}

\subsection{Named Entities} \label{subsec:entity}
Named entities (NEs) are an essential part of sentences in terms of human understanding and readability. The mistranslation of NEs can significantly impact translation output, although evaluation scores (e.g. \BLEU{}) may not be adversely affected. 
Therefore, we also annotate named entities in the reference documents, following a similar procedure to OntoNotes~\cite{weischedel2013ontonotes}.  In total, 2,234 entities are annotated in the \BWB{} test.

\subsection{Pronouns} \label{subsec:pronoun}
Pronoun translation has been the focus of discourse-level MT evaluation~\cite{hardmeier-2012-discourse, APT-2017}.
As showned in \Cref{tab:pronoun}, there are significantly fewer pronouns in Chinese due to its pronoun-dropping property.
This poses extra challenges for NMT since the skill of anaphoric resolution is required. 

\section{Experiments} \label{sec:exp}
We carry out evaluation of both baseline and state-of-the-art MT models on \BWB{} and also provide human post-editing performance \PE{} for comparison.
The following 6 baselines are adapted:\footnote{Professional translators were hired to conduct post-editing on the \OMTc{} outputs. They were instructed to correct only discourse-level errors with minimal modification. 
\MTS{} and \MTD{} are trained on \BWB{} by fairseq~\cite{fairseq}, and the training details are in App.\ref{app:model_parameters}.}
\begin{itemize}
    \vspace{-5pt}
    \item \bsc{\SMT{}}: phrase-based baseline~\cite{SMT}.
    \vspace{-8pt}
    \item \bsc{\OMTc}, \bsc{g}oo\bsc{gl}e, \bsc{b}ai\bsc{d}u: commercial systems.
     \vspace{-8pt}
    \item \bsc{\MTS}: the Transformer baseline that translates sentence by sentence \cite{transformer}.
     \vspace{-8pt}
    \item \bsc{\MTD}: the document-level NMT model that adopts two-stage training \cite{ctx}.
     \vspace{-8pt}
\end{itemize}

\paragraph{Evaluation Metrics}
Systems are evaluated with automatic standard sentence-level MT metrics (\BLEU{}~\cite{BLEU}, \METEOR~\cite{Meteor}, \BERTScore~\cite{BERTScore}) and a document-level metric~\citep[\BlonD,][]{blonde}. 
We also performed evaluation targeted at specific discourse-phenomena.

\begin{figure}[t]
    \centering
    \vspace{-5pt}
    \includegraphics[width=0.45\textwidth,trim={0cm 7.5cm 1.5cm 0.5cm} ,clip]{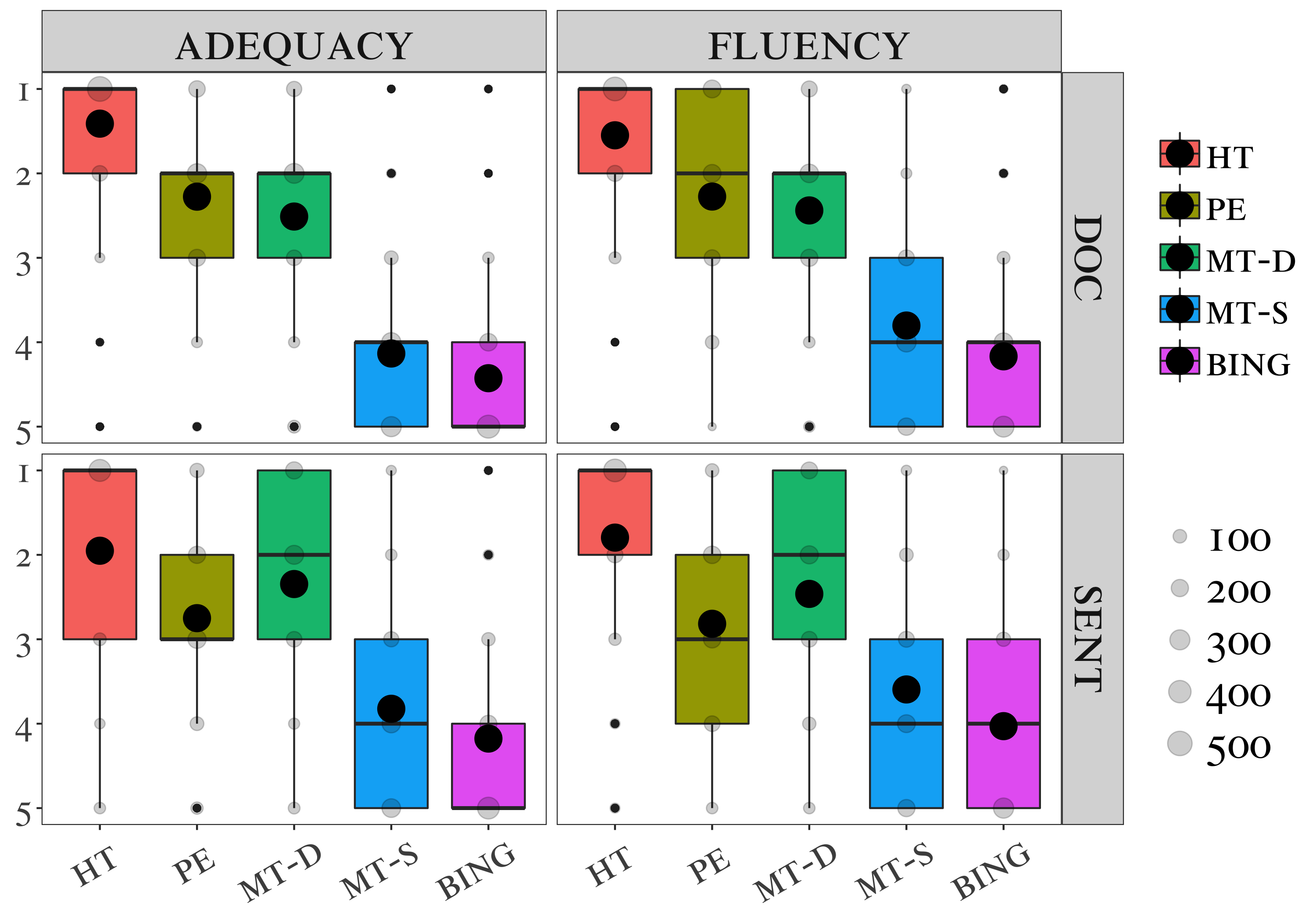}
    \vspace{-10pt}
    \caption{Human evaluation results on \BWB{}. Each $\bullet$ represents the average of the system's rankings. \looseness=-1}
    \vspace{-14pt}
    \label{fig:human_evaluation}
\end{figure}

\paragraph{Human Evaluation}
The human evaluation on \BWB{} is conducted by 8 professional asselssors, i.e. 4 Chinese to English translators and 4 native English revisers. Following the recommendations of \citet{laubli-etal-2020-set}, we evaluate two units of linguistic context (\Sentence{} and \Document{}) independently based on their respective \fluency{} and \adequacy{}.
Spam items are used for quality control \cite{kittur2008crowdsourcing}.\footnote{App. \ref{app:human_evaluation} describes how human assessment is carried out. The inter-rater agreement is reported in \Cref{tab:kappa}.}
The results and rankings of the aforementioned systems and human translation (\HT) are listed in \Cref{fig:human_evaluation}.
The large gap between the performances of \HT{} and \MT{} indicates that the genre of \BWB{}, i.e., literary translation, is very challenging for \MT{}, and NMT systems are far from human parity.
\MTD{} performs significantly better than \MTS{}, suggesting that \BWB{} contains rich discourse phenomena that can only be translated accurately when the context is taken into account.
It is also worth noting that even though \PE{} is the post-editing of the poorly performing system \OMTc{}, it is still surprisingly able to achieve better performance than \MTD{} at the document level.
This observation confirms the claim that the discourse phenomena contained in \BWB{} have a huge impact on human judgment of translation quality.

\section{Related Work} \label{sec:related_work}
\paragraph{Document-Level Parallel Corpora}
There are some document-level parallel corpora in the market: TED Talks of IWSLT dataset~\cite{iwslt2020}, News Commentary~\cite{tiedemann-2012-OPUS}, LDC\footnote{\url{https://www.ldc.upenn.edu}} and OpenSubtitle~\cite{opensubtitles2018}. The sizes and average length of these corpora are summarized in \Cref{tab:datasets}. Detailed descriptions of these corpora are in \Cref{app:dataset}. \BWB{} is the largest corpus in terms of size. Moreover, the sentences and documents in \BWB{} are substantially longer than previous corpora. It is also worth noting that \BWB{} differs from previous corpora in terms of genre.

\paragraph{Evaluation Test Suites for Document-Level MT} 
Evaluating document-level translation quality is difficult for metrics such as \BLEU{}.  
Therefore, many test suites that perform context-aware evaluation have been proposed~\citep[][\textit{inter alia}]{DiscoMT, PROTEST, burchardt-etal-2017-linguistic, isabelle-etal-2017-challenge, rios-gonzales-etal-2017-improving, muller-etal-2018-large,bawden-etal-2018-evaluating,voita-etal-2019-good, guillou-hardmeier-2018-automatic}. However, the scope of most test suites has been restricted to pronouns and limited in size. 
In contrast, \BWB{} annotates not only pronouns but also other context-sensitive spans that \textit{cannot} be translated correctly by context-agnostic systems. 

\section{Conclusion}
We presented a newly constructed document level parallel corpus \BWB{} with annotations of various discourse phenomena such as discourse-sensitive spans and named entities in the test set to allow for a fine-grained analysis of document level translation. 
Experiments shows that \BWB{} is challenging for existing NMT models and could serve as a good benchmark for coherent document level translation.

\section*{Limitations}
As of now, this corpus consists of only Chinese-to-English data. 
In addition, as illustrated in \Cref{app:case_study}, coreference resolution is also crucial to document level machine translation. We have left coreference annotation to future work.

\section*{Ethical Considerations}
The annotators were paid a fair wage and the annotation process did not solicit any sensitive information from the annotators. 
In regard to the copyright of our dataset, as stated in the paper, the crawling script that we plan to release will allow others to reproduce our dataset faithfully and will not be in breach of any copyright. 
In addition, the release of our annotated test set will not violate the doctrine of \textbf{Fair Use} (US/EU), as the purpose and character of the use is \emph{transformative}.
Please refer to \url{https://www.nolo.com/legal-encyclopedia/fair-use-the-four-factors.html} for relevant laws.


\bibliography{mine}
\bibliographystyle{acl_natbib.bst}

\clearpage
\begin{appendices}
\section{Dataset Statistics} \label{app:dataset}
\subsection{Existing Corpora}
Here we review the existing document-level parallel corpora mentioned in \Cref{sec:related_work} in detail. The full list of their statistics is provided in \Cref{tab:datasets_full}.
\paragraph{LDC}
This corpus consists of formal articles from the news and law domains.
The articles ave syntactic structures such as conjoined phrases, which make machine translation challenging.
However, the news articles in this corpus are relatively outdated.

\paragraph{IWSLT}
This corpus contains the TED Talks that covers the variety of topics. However, it is quite small in scale, which makes training large transformer-based models impractical.

\paragraph{News Commentary} 
This corpus consists of political and economic commentary crawled from the web site Project Syndicate\footnote{\url{4https://wit3.fbk.eu}}.  However, the scale of this corpus is also quite small. Moreover, there are no parallel Chinese-to-English data available in this corpus.

\paragraph{Europarl}
The corpus is extracted from the proceedings of the European Parliament. Only European language pairs are available in this corpus.

\paragraph{OpenSubtitle}
This corpus is a collection of translated movie
subtitles \cite{lison-tiedemann-2016-opensubtitles2016}. Besides being simple and short, the ``documents'' in this corpus are usually verbal and informal as well.

\subsection{Statistics of \BWB{}}
The statistics of the training, development and test sets of \BWB{} is provided in \Cref{tab:dataset_split}.

\begin{table*}[htpb]
\begin{adjustbox}{width=0.99\textwidth}
{
\centering
\begin{tabular}{l|c|c|rrr|rrr}
\toprule[2pt]
\multirow{2}{*}{Corpus}  & \multirow{2}{*}{Genre}   & \multirow{2}{*}{Language} & \multicolumn{3}{c}{Size}                                                                         & \multicolumn{3}{c}{Averaged Length}                                                             \\
                         &                           &                           & \#word & \#sent & \#doc & \#word/sent & \#sent/doc & \#word/doc \\
\midrule[1pt]
	
\rowcolor{LightCyan}  \multirow{4}{*}{IWSLT~\cite{iwslt2020}}   & TED talk & ZH-EN                     & 4.2M                         & 0.2M                            & 2K                              & 19.5                           & 100                            & 2,100                         \\
                         &     TED talk        & FR-EN                     & 4.4M                         & 0.2M                            & 2K                              & 20.8                           & 100                            & 2,190                         \\
                         &    TED talk      & ES-EN                     & 4.2M                         & 0.2M                            & 2K                              & 19.9                           & 100                            & 2,080                         \\
                         &      TED talk     & DE-EN                     & 4.1M                         & 0.2M                            & 2K                              & 19.3                           & 100                            & 2,070                         \\
\midrule[1pt]
\multirow{2}{*}{NewsCom~\cite{tiedemann-2012-OPUS}} & News     & ES-EN                     & 6.4M                         & 0.2M                            & 5K                              & 30.7                           & 40                             & 1,288                         \\
                         &      News     & DE-EN                     & 6.4M                         & 0.2M                            & 5K                              & 33.1                           & 40                             & 1,288                         \\
\midrule[1pt]
Europarl~\cite{koehn-2005-europarl}                 & Parliament                & ET-EN                     & 7.3M                         & 0.2M                            & 15K                             & 35.1                           & 13                             & 485                           \\
\midrule[1pt]
\rowcolor{LightCyan}  LDC                     & News                      & ZH-EN                     & 81.8M                        & 2.8M                            & 61K                             & 23.7                           & 46                             & 1,340                         \\
\midrule[1pt]
\multirow{3}{*}{OpenSub~\cite{opensubtitles2018}} & Subtitle & FR-EN                     & 219.0M                         & 29.2M                           & 35K                             & 8.0                              & 834                            & 6,257                         \\
                         &   Subtitle     & EN-RU                     & 183.6M                       & 27.4M                           & 35K                             & 5.8                            & 783                            & 5,245                         \\
\rowcolor{LightCyan}             &      Subtitle     & ZH-EN                     & 16.9M                        & 2.2M                            & 3K                              & 5.6                            & 733                            & 5,647                         \\
\midrule[1.5pt]
\rowcolor{LightCyan}   \BWB{} (chapter)    &     Novel       & ZH-EN                     & \textbf{460.8M}                       & \textbf{9.6M}                            & \textbf{196K}                            & \textbf{48.1}                          & 49                             & 2,356                         \\
 \rowcolor{LightCyan} \BWB{} (book) &    Novel           & ZH-EN                     & \textbf{460.8M}                       & \textbf{9.6M}                            & 384                             & \textbf{48.1}                           & \textbf{25.0K}                          & \textbf{1.2M}          \\
\bottomrule[2pt]
\end{tabular}
}
\end{adjustbox}
    \caption{Statistics of various document-level parallel corpora. The parallel Chinese-English data is highlighted in \colorbox{LightCyan}{Cyan}.}
    \label{tab:datasets_full}
\end{table*}

\begin{table*}[htpb]
\small
\centering
\begin{tabular}{c|cccc|cccc}
\toprule[2pt]
\multirow{2}{*}{Split}   & \multicolumn{4}{c}{Size}  & \multicolumn{3}{c}{Averaged Length}  \\
& \#word & \#sent & \#chap & \#book &  \#word/sent & \#sent/chap & \#chap/book \\
\midrule[1pt]
Train & 325.4M & 9.57M                     & 196K                   & 378                        & 34.0                           & 48.8                            & 519.3                           \\
Valid   & 68.0K  & 2,632 & 80 & 6                          & 25.8                           & 32.9                            & 13.3                            \\
Test  & 67.4K  & 2,618 & 79 & 6                          & 25.7                           & 33.1                            & 13.2   \\
\bottomrule[2pt]
\end{tabular}
    \caption{Statistics of the training, development and test sets of \BWB{}.}
    \label{tab:dataset_split}
\end{table*}

\section{Case Study} \label{app:case_study}
We provide two example chapters in \BWB{} with coreference annotation in \Cref{fig:example_book1_0} and  \Cref{fig:example_book153_0}. 
We observe that the \BWB{} dataset poses challenges for NMT in the following ways.

\paragraph{Entity Consistency}
There are many named entities in the dataset that have a high repetition rate, such as fictional characters. 
Therefore, named entity consistency is a significant challenge in machine translation on this dataset.
For example, the translations of \orangeEnt{Weibo} and \redEnt{Qiao Lian} in \Cref{fig:example_book1_0} are not consistent in context.

\paragraph{Entity Recognition and Retrieval}
In addition to the fluency of entity translation, the adequacy of entity translation is another challenge in \BWB{}.
In the case of fictional characters with strange names, the NMT model may not correctly \emph{recognize} named entities, resulting in extremely poor translation quality, as in \Cref{fig:example_book153_0}. ``Ye Qing Luo'' could be literally translated as ``night'', ``clear'', ``fall''; however, it is actually a fictional characters.

Even though fictional characters are difficult to translate, they are relatively rare throughout the text, so it would be beneficial to abandon the assumption of inter-sentence independence in consideration of global contextual information.
One potential way to alleviate this problem is to equip NMT models with an entity recognition module.

\paragraph{Anaphoric Information Loss}
Chinese, being one of the pro-drop languages, omits many pronouns, while the English language does not, as shown in \Cref{tab:pronoun}. 
Translating from Chinese to English thus requires context to infer the correct English pronouns to compensate for the anaphoric information loss of sentence-level Chinese-to-English translations.

\paragraph{Morphological Information Loss}
Tense information is also frequently absent in Chinese and can only be inferred from context.
In general, this problem, which we refer to as \emph{morphological information loss}, is often encountered when translating from a morphologically poorer language to a morphologically richer one.
In the case of Chinese-to-English translation, tense information is often lost, while in other language pairs, such as English-to-French and English-to-German, gender information is often missed since as French and German are morphologically richer than English.

\paragraph{Coreference}
In addition, in \Cref{fig:example_book1_0}, we observe that the focus entity of the document is shifting throughout the text (\redEnt{Qiao Lian} $\xrightarrow{}$ \bleuEnt{Shen Liangchuan} $\xrightarrow{}$ \pinkEnt{Wang Wenhao} $\xrightarrow{}$ \bleuEnt{Shen Liangchuan} $\xrightarrow{}$ \cyanEnt{Song Cheng}), and this information is language-independent, i.e. consistent in source and target. This information could be used to improve the coherence of translation.


\section{Experiment Setup} \label{app:model_parameters}
We adopt the parameters of Transformer Big \cite{transformer} for both \MTS\ and \MTD.
More precisely, the layers in the big encoders and decoders are $N=12$ , the number of heads per layer is $h = 16$, the dimensionality of input and output is $d_{model} = 1024$, and the inner-layer of a feed-forward networks has dimensionality $d_{ff} = 4096$. The dropout rate is fixed as 0.3. We adopt Adam optimizer with $\beta_1 = 0.9, \beta_2 = 0.98, \epsilon = 10^{-9}$, and set learning rate $0.1$ of the same learning rate schedule as Transformer. We set the batch size as 6,000 and the update frequency as 16 for updating parameters to imitate 128 GPUs on a machine with 8 V100 GPU. The datasets are encoded by BPE with 60K merge operations.

\section{Human Evaluation} \label{app:human_evaluation}
We conducted human evaluation on the \BWB{} test set following the protocol proposed by \citep{laubli-etal-2018-machine, laubli-etal-2020-set}. 
As stated in \Cref{sec:exp}, we evaluated two units of linguistic context (\Sentence{} and \Document{}) independently based on their respective \fluency{} and \adequacy{}.
We showed raters isolated sentences in random order in the \Sentence-level evaluation, whereas in the \Document-level evaluation, we presented entire documents and asked raters to evaluate a sequence of five sequential sentences at a time in order. 
The \adequacy{} evaluation was based solely on source texts, whereas neither source texts nor references were included in the \fluency{} evaluation.

The \adequacy{} evaluation was conducted by four professional Chinese to English translators, and the \fluency{} evaluation was conducted by four native English revisers.
The four translators were different from the professional translators who performed human translation. For human evaluation, we deliberately invite another group of specialists to avoid making judgments biased towards human translation. 

We adopted relative ranking because it has been shown to be more effective than direct assessment when conducted by experts rather than crowd workers~\cite{barrault-etal-2019-findings}.
In particular, raters were presented with the system outputs and were asked to evaluate the system outputs vis-à-vis one another, e.g. to decide whether system A was better than system B (with ties allowed). 

By randomizing the order of presentation of the system outputs, we were able to blind the origin of the output sentences and documents. While in the \Sentence-level evaluation, the system outputs were presented in different orders for each sentence, the \Document-level evaluation used the same ordering of systems within a document to help raters better assess global coherence.

Additionally, we used spam items for quality control.\cite{kittur2008crowdsourcing}. 
At the \Sentence-level, we make one of the five options nonsensical in a small fraction of items by randomly shuffling the order of the translated words, except for 10\% at the beginning and end. 
At the \Document-level, we randomly shuffle all translated sentences except the first and last sentence at the document level, rendering one of the five options nonsensical. 
If a rater marks a spam item as better than or equal to an actual translation, this is a strong indication that they did not read both options carefully.

Each raters evaluated 180 documents (including 18 spam items) and 180 sentences (including 18 spam items). The 180 sentences were randomly sampled from \testset{1} or \testset{2}.
We spited the test set into two non-overlapping subsets, referred to as \testset{1} and \testset{2}.
Note that \testset{1} and  \testset{2} were chosen from different books. 
Each rater evaluated both sentences and documents, but never the same text in both conditions so as to avoid repetition priming \cite{gonzalez2011cognitive}. 
Each document or sentence was therefore evaluated by two raters, as shown in \Cref{tab:human_evaluation_units}.

We report pairwise inter-rater agreement in \Cref{tab:kappa}. Cohen's kappa coefficients were used: 
\begin{equation}
    \kappa = \frac{P(A)-P(E)}{1-P(E)}
\end{equation}
where $P(A)$ is the proportion of times that two raters agree, and $P(E)$ is the likelihood of agreement by chance.

\begin{table}[]
\centering
\begin{tabular}{c|cc|cc}
\toprule[2pt]
 & \multicolumn{2}{c|}{\testset{1}} & \multicolumn{2}{c}{\testset{2}} \\
\adequacy  & \textsc{sent} & \textsc{doc}   & \textsc{sent} & \textsc{doc} \\ 
\midrule[1pt]
\rater{1} &             & \checkmark & \checkmark &             \\
\rater{2} &             & \checkmark & \checkmark &             \\
\rater{3} & \checkmark  &            &            & \checkmark  \\
\rater{4} & \checkmark  &            &            & \checkmark  \\ 
\bottomrule[2pt]
\end{tabular}
\begin{tabular}{c|cc|cc}
\toprule[2pt]
 & \multicolumn{2}{c|}{\testset{1}} & \multicolumn{2}{c}{\testset{2}} \\
\fluency  & \textsc{sent} & \textsc{doc}   & \textsc{sent} & \textsc{doc} \\ 
\midrule[1pt]
\rater{5} &             & \checkmark & \checkmark &             \\
\rater{6} &             & \checkmark & \checkmark &             \\
\rater{7} & \checkmark  &            &            & \checkmark  \\
\rater{8} & \checkmark  &            &            & \checkmark  \\ 
\bottomrule[2pt]
\end{tabular}
\caption{The evaluation units and corresponding raters. \rater{1-4} are professional Chinese to English translators and \rater{5-8} are native English revisers. }
\label{tab:human_evaluation_units}
\end{table} 
 
\begin{table}[]
\centering
\begin{tabular}{ccc}
\toprule[2pt]
              & \Sentence & \Document  \\ 
\midrule[1pt]
\rater{1}-\rater{2} & .171 & .169 \\
\rater{3}-\rater{4} & .294 & .346 \\
\rater{5}-\rater{6} & .323 & .402 \\
\rater{7}-\rater{8} & .378 & .342 \\ 
\bottomrule[2pt]
\end{tabular}
\caption{Inter-rater agreements measure by Cohen's $\kappa$.}
\label{tab:kappa}
\end{table}
 
\begin{figure*}[t]
\begin{adjustbox}{width=\textwidth}
\centering \scriptsize
\begin{tabular}{p{5pt}p{150pt}p{150pt}p{150pt}}
\toprule[2pt]
& SOURCE & REFERENCE & MT \\
\midrule[1pt]
1) & \ZH{	\redEnt{乔恋}攥紧了拳头，垂下了头。	}&	\redEnt{Qiao Lian} clenched \redEnt{her} fists and lowered \redEnt{her} head.	&	\redEnt{Joe} clenched \redEnt{his} fist and bowed \redEnt{his} head.	\\
2) & \ZH{	其实\bleuEnt{他}说得对。	}&	Actually, \bleuEnt{he} \Verb{was} right.	&	In fact, \bleuEnt{he}\Verb{'s} right.	\\
3) & \ZH{自己就是一个蠢货，竟然会相信了网络上的爱情。	}&	\redEntOmit{She} \Verb{was} indeed an idiot, as only an idiot \Verb{would} believe that they could find true love online.	&	\redEntOmit{I} \Verb{am} a fool, even \Verb{will} believe the love on the Internet.	\\
4) & \ZH{	\redEnt{她}勾起了嘴唇，深呼吸一下，正打算将手机放下，微信上却被炸开了锅。	}&	\redEnt{She} curled \redEntOmit{her} lips and took a deep breath. Just when \redEntOmit{she} was about to put down \redEntOmit{her} cell phone, a barrage of posts bombarded \redEntOmit{her} WeChat account.	&	\redEnt{She} ticked \redEntOmit{her} lips, took a deep breath, and was about to put \redEntOmit{her} phone down, but weChat was blown open.	\\
5) & \ZH{	\redEnt{她}点进去，发现是\orangeEnt{凉粉群}，所有人都在@\redEnt{她}。	}&	\redEnt{She} logged into \redEntOmit{her} account and saw that a large number of fans in the \orangeEnt{Shen Liangchuan fan group} had tagged \redEnt{her}.	&	She nodded in and found it was a \orangeEnt{cold powder group}, and everyone was on \redEnt{her}.	\\
6) & \ZH{	【\redEnt{乔恋}：怎么了？	}&	[\redEnt{Qiao Lian}: What happened?]	&	\redEnt{Joe}: What's the matter?	\\
7) & \ZH{	【\grayEnt{川流不息}：\redEnt{乔恋}，快看\orangeEnt{微博}头条！ \orangeEnt{微博}头条？	}&	[\grayEnt{Chuan Forever}: \redEnt{Qiao Lian}, look at the headlines on \orangeEnt{Weibo}, quickly!]	&	\grayEnt{Chuan-flowing}: \redEnt{Joe love}, quickly look at the \orangeEnt{micro-blogging} headlines! \orangeEnt{Weibo} headlines?	\\
8) & \ZH{	\redEnt{她}微微一愣，拿起手机，登陆\orangeEnt{微博}，在看到头条的时候，整个人一下子愣住了！	}&	\redEnt{She} froze momentarily, then picked up \redEntOmit{her} cell phone and logged into \orangeEnt{Weibo}. When  \redEntOmit{she} saw the headlines,  \redEntOmit{her entire body} immediately froze over again!	&	\redEnt{She} took a slight look, picked up the phone, landed on the \orangeEnt{micro-blog}, when  \redEntOmit{she} saw the headlines,  \redEntOmit{the whole person} suddenly choked!	\\
9) & \ZH{	剧组发布会。 \bleuEnt{沈凉川}应邀出场，\grayEnt{导演}立马恭敬地迎接过来，客气的跟\bleuEnt{他}说这话，表达着\grayEnt{自己}对\bleuEnt{他}能够到来的谢意。	}&	\bleuEnt{Shen Liangchuan} arrived at the scene after accepting the invitation. \grayEnt{The director} immediately went to greet \bleuEnt{him} in a respectful manner, politely welcoming \bleuEnt{him} and expressing \grayEnt{his} gratitude for \bleuEnt{Shen Liangchuan}’s presence today.	&	The show's release. \bleuEnt{Shen Liangchuan} was invited to appear, \grayEnt{the director} immediately greeted \bleuEnt{him} with respect, politely said this to \bleuEnt{him}, expressed \grayEnt{his} gratitude for \bleuEnt{his} arrival.	\\
10) & \ZH{	对\bleuEnt{沈凉川}没有说话，看向不远处的\pinkEnt{王文豪}。	}&	\bleuEnt{Shen Liangchuan} did not speak. Instead \bleuEntOmit{he} looked at \pinkEnt{Wang Wenhao}, who was nearby.	&	\bleuEnt{Shen Liangchuan} did not speak, look not far from \pinkEnt{Wang Wenhao}.	\\
11) & \ZH{	\pinkEnt{王文豪}出事以后，所有的作品全部下架，而这一部剧还能播出，是因为\pinkEnt{王文豪}在里面友情饰演的男三号戏份很少，几乎可以忽略不计。	}&	After \pinkEnt{Wang Wenhao}’s scandal broke, every film \pinkEntOmit{he} starred in had been taken down. Only this show \Verb{could} still be broadcasted, as \pinkEnt{Wang Wenhao} \Verb{had} a supporting role in it and \Verb{was} practically unnoticeable.	&	After \pinkEnt{Wang Wenhao}'s accident, all the works were off the shelves, and this play \Verb{can} also be broadcast, because \pinkEnt{Wang Wenhao} in the friendship played by the male no. 3 play \Verb{is} very few, almost negligible.	\\
12) & \ZH{	剧组根本就没有邀请\pinkEnt{王文豪}，可\pinkEnt{他}却不知道从哪里拿到了邀请函，自己堂而皇之的进来了。 \pinkEnt{他}当然要进来了。	}&	In fact, the cast and crew hadn’t even invited \pinkEnt{Wang Wenhao}. However, \pinkEnt{he} had obtained a copy of the invitation letter somehow, and \Verb{strode} imposingly into the venue anyway.	&	The crew did not invite \pinkEnt{Wang Wenhao}, but \pinkEnt{he} did not know where to get the invitation, \pinkEnt{his} own entrance. Of course \pinkEnt{he}\Verb{'s} coming in.	\\
13) & \ZH{	这是\pinkEnt{他}最后的机会了。	}&	After all, this \Verb{was} \pinkEnt{his} final chance.	&	This \Verb{is} \pinkEnt{his} last chance.	\\
14) & \ZH{	丑闻闹出来，几乎所有的广告商和剧组都跟他毁约。	}&	After \pinkEntOmit{his} scandals broke, practically every advertiser and filming crew wanted to break their contracts with \pinkEnt{him}.	&	The scandal broke, and almost all advertisers and crews broke \pinkEntOmit{his} contract with \pinkEnt{him}.	\\
15) & \ZH{	\pinkEnt{他}现在宁可拍男三号，也不想就此沉寂。	}&	\pinkEnt{He} would rather take a supporting role than fade out into obscurity.	&	\pinkEnt{He} would rather shoot the men's number three now than be silent about it.	\\
16) & \ZH{	因为\pinkEnt{他}的事情，根本就压不下去。	}&	That was because the scandals surrounding \pinkEnt{him} \Verb{would} never disappear.	&	Because of \pinkEnt{his} affairs, there \Verb{is} no pressure.	\\
17) & \ZH{	所以\pinkEnt{王文豪}在发布会上，到处讨好别人。	}&	Thus, \pinkEnt{Wang Wenhao} went around trying to curry favor with everybody at this press conference.	&	So \pinkEnt{Wang Wenhao} tried to please others at the press conference.	\\
18) & \ZH{	\bleuEnt{沈凉川}穿着一身深灰色西装，面色清冷，手里端着一个高脚香槟杯，站在桌子旁边，整个人显得格外俊逸，却也格外的清冷，让周围的人都不敢上前搭讪。	}&	\bleuEnt{Shen Liangchuan} was wearing a dark grey suit and \bleuEntOmit{he} had a cold expression. \bleuEntOmit{He} was holding a champagne glass and was currently standing beside a table. \bleuEntOmit{He} looked exceptionally stylish, but also exceptionally icy. As a result, none of the people around \bleuEntOmit{him} dared to approach \bleuEntOmit{him}.	&	\bleuEnt{Shen 
River} was wearing a dark gray suit, \bleuEntOmit{his} face was cold, and \bleuEntOmit{he} was holding a tall champagne glass in \bleuEntOmit{his} hand. Standing beside the table, the whole person looked extraordinarily handsome, but also extraordinarily cold, so that people around \bleuEntOmit{him} did not dare to approach \bleuEntOmit{him}. 	\\
19) & \ZH{	可如果能注意到\bleuEnt{他}，就会发现\bleuEnt{他}的视线，却总是若有似无的飘到\pinkEnt{王文豪}身上。	}&	If anyone \Verb{had} paid attention to \bleuEnt{him}, they \Verb{would} have noticed that \bleuEnt{his} gaze \Verb{kept} drifting over to \pinkEnt{Wang Wenhao}.	&	But if you \Verb{can} notice \bleuEnt{him}, you \Verb{will} find \bleuEnt{his} sight, but always if there \Verb{is} nothing floating to \pinkEnt{Wang Wenhao} body.	\\
20) & \ZH{	\cyanEnt{宋城}站在\bleuEnt{他}的身边，察觉到这一点以后，就忍不住拽了拽\bleuEnt{他}的胳膊。	}&	\cyanEnt{Song Cheng} stood at \bleuEnt{his} side. After noticing \bleuEntOmit{his} behavior, \cyanEntOmit{he} \Verb{could} not help but pinch \bleuEnt{his} arm.	&	\cyanEnt{Songcheng} stood by his side, aware of this, \Verb{can} not help but pull \bleuEnt{his} arm.	\\
21) & \ZH{	\bleuEnt{沈凉川}淡淡回头，看向\cyanEnt{他}，目露询问。	}&	\bleuEnt{Shen Liangchuan} turned around and looked at \cyanEnt{him} casually, with a questioning face.	&	\bleuEnt{Shen Liangchuan} faint lying back, looked at \cyanEnt{him}, blind inquiry.	\\
22) & \ZH{	“\bleuEnt{沈哥}，您到底是要干什么啊？ 能不能告诉我，好让我有个心理准备。 您这样突然跑过来参加这么一个小剧组的发布会，又什么都不说就这么杵着，我心里瘆的慌。”	}&	“\bleuEnt{Brother Shen}, what are you planning to do? Can you tell me beforehand so that I can prepare myself mentally. You suddenly decide to come and attend such a small-scale press conference, yet you have been completely silent and are now just standing here and doing nothing? My heart is beating anxiously right now.”	&	\bleuEnt{Shen brother}, what the hell are you doing? Can you tell me so that I have a mental preparation. You suddenly ran over to attend the launch of such a small group, and said nothing so, I panicked. 	\\
23) & \ZH{	\bleuEnt{沈凉川}听到这话，抿了一口香槟，接着，将香槟杯放下。	}&	After \bleuEnt{Shen Liangchuan} heard him speak, he sipped a mouthful of champagne and put the glass down.	&	\bleuEnt{Shen} Said, took a sip of champagne, and then put the champagne glass down.	\\
24) & \ZH{	旋即，他迈开了修长的步伐。	}&	Then, \bleuEnt{he} walked away in long strides.	&	Immediately, \bleuEnt{he} took a slender step.	\\
25) & \ZH{	\cyanEnt{宋城}的心都提了起来，紧跟在\bleuEnt{他}身后。 \bleuEnt{沈凉川}一步一步往前，走到了前方。	}&	\cyanEnt{Song Cheng} was extremely nervous and followed \bleuEnt{him}. \bleuEnt{Shen} Liangchuan walked forward, one step at a time, until \bleuEntOmit{he} reached the front of the room.	&	\cyanEnt{Song Cheng}'s heart was raised and followed immediately behind \bleuEnt{him}. \bleuEnt{Shen Liangchuan} step by step forward, walked forward.	\\
26) & \ZH{	\pinkEnt{王文豪}正在跟别的三流小明星套近乎，那个人询问，“听说你打了一个狗仔？”	}&	\pinkEnt{Wang Wenhao} \Verb{was} currently ingratiating himself with a C-list celebrity. The celebrity asked, “Hey, I heard that you beat a paparazzi?”	&	\pinkEnt{Wang Wenhao} \Verb{is} with other third-rate star-studded sets, the man asked, "I heard you hit a paparazzi?" "	\\
27) & \ZH{	“对啊，现在的狗仔就是惹人厌恶，\pinkEnt{我}早就想动手教训他们了！ “\pinkEnt{你}这样，就不怕跟他们结仇啊？”	}&	“Yeah, the paparazzi nowadays are so disgusting. \pinkEnt{I} have wanted to teach them a lesson myself for some time now!” "Are not \pinkEnt{you} afraid of becoming an enemy of them?"	&	"Yeah, the paparazzi now are disgusting, \pinkEnt{I} have long wanted to teach them a lesson! "If \pinkEnt{you} are like this, \pinkEnt{you} are not afraid of enmity with them? "	\\
28) & \ZH{	''\pinkEnt{我}都这样了，\pinkEnt{我}怕什么？ 当初\bleuEnt{沈影帝}以正当防卫为借口，将一名狗仔打了，告到了法庭上去不也不了了之吗？``	}&	“\pinkEnt{I}’ve already done it, so what should \pinkEnt{I} be scared of? That time \bleuEnt{Best Actor Shen} beat up a reporter, \bleuEntOmit{he} claimed that it was in self-defence so that \bleuEntOmit{he} would have an excuse if he got sued, right? At that time, nobody said anything” 	&	"\pinkEnt{I}'m already like this, what am \pinkEnt{I} afraid of? Back then, \bleuEnt{Yingdi Shen} beat up a paparazzi under the pretext of self-defense, and sued \bleuEnt{him} in court, wouldn't it be over?	\\
29) & \ZH{	\pinkEnt{王文豪}说到这里，嘿嘿一笑。 还想说什么，忽然察觉到身后有人靠近。	}&	As \pinkEnt{Wang Wenhao} spoke, \pinkEntOmit{he} laughed heartily. Just as \pinkEntOmit{he} was about to continue speaking, \pinkEntOmit{he} suddenly felt a presence approaching \pinkEntOmit{him} from behind.	&	When \pinkEnt{Wang Wenhao} said this, \pinkEntOmit{he} smiled. \pinkEntOmit{He} was about to say something, but suddenly noticed someone approaching behind \pinkEntOmit{him}.	\\
30) & \ZH{	\pinkEnt{他}回头，就看到\bleuEntOmit{沈凉川}，眼瞳一缩，舔着脸笑，却见\bleuEntOmit{沈凉川}往前一步，一把揪住\pinkEnt{他}的衣领，接着一拳头对着\pinkEnt{他}的脸就砸过来！	}&	He turned around and saw \bleuEnt{Shen Liangchuan}. \bleuEntOmit{His} eyes narrowed and attempted to smile at \pinkEntOmit{him}. However, \bleuEnt{Shen Liangchuan} took a step forward, grabbed \pinkEnt{his} tie and threw a punch at \pinkEnt{his} face!	&	\pinkEnt{He} looked back, \pinkEntOmit{he} saw \bleuEnt{Shen Liangchuan}, eyes shrink, licking \bleuEntOmit{his} face and smiling, but saw Shen Liangchuan a step forward, a holding \pinkEnt{his} collar, and then a fist to \pinkEnt{his} face hit!	\\
\bottomrule[2pt]
\end{tabular}
\end{adjustbox}
\caption{An example chapter in \BWB. The same entities are marked with the same color. 
Pronoun omissions are marked with \Omit{}.
The mistranslated verbs are marked with \Verb{teal},
and the mistranslated named entities are marked with \orangeEnt{}.}
\label{fig:example_book1_0}
\vspace{-5pt}
\end{figure*}

\begin{figure*}[t]
\begin{adjustbox}{width=\textwidth}
\centering \scriptsize
\begin{tabular}{p{5pt}p{150pt}p{150pt}p{150pt}}
\toprule[2pt]
& SOURCE & REFERENCE & MT \\
\midrule[1pt]
1) & \ZH{	\redEnt{夜清落}浑身上下都传来剧烈的疼痛感，宛如千万把利刃，切割着\redEnt{她}的身体。	}&	\redEnt{Ye Qing Luo} suddenly felt an excruciating sharp pain tormenting \redEntOmit{her} entire body. It seemed as if a million sharp blades were slashing at \redEnt{her}.	&	The night fell all over the body came a sharp pain, like a thousand sharp blades, cutting \redEntOmit{her} body.	\\
2) & \ZH{	尤其心脏那处，像是有着一团烈火，在体内燃烧，肆意的烧灼着\redEnt{她}的一切。	}&	\redEntOmit{Her} heart felt as if it was burning and that flame threatened to burn everything.	&	Especially the heart, like a fire, burning in the body, burning \redEnt{her} everything.	\\
3) & \ZH{	\redEnt{夜清落}想要伸手揉揉眉心，却发现自己全身虚软无力，连手指头都没法动弹。	}&	\redEnt{Ye Qing Luo} wanted to reach out but \redEntOmit{she} found that \redEntOmit{she} couldn’t move. \redEntOmit{She} felt so weak that \redEntOmit{she} could not even lift a finger.	&	\redEnt{Night clear} wants to reach out to rub the eyebrows, but found that \redEntOmit{their} whole body is weak, even fingers can not move.	\\
4) & \ZH{	耳边，隐隐约约传来模糊不清的声音。 “四少爷，这么做真的没事吗？” 	}&	In \redEntOmit{her} ears, there was a vague sound. “Fourth Young Master, are you sure it’s really alright to do this?”	&	in my ear, vaguely came a vague sound. "Four Masters, is it really okay to do this?"	\\
5) & \ZH{	“怕什么！”	}&	What are you so afraid of!	&	"Afraid of what! "	\\
6) & \ZH{	“太子殿下怎么可能看得上这个废物？ 要不是看她担着三泉宗少宗主的名头，太子殿下早就将她踹了。”	}&	How can His Royal Highness put such a good-for-nothing waste in his eyes? If not for her status as the Three Spring’s Lord’s daughter, do you think he would even bother with her?	&	"How could His Highness see this waste? Had it not been for her bearing the name of the Lord of Sanquan Zong, His Royal Highness would have taken her. "	\\
7) & \ZH{	“小废物可是夜四小姐亲自送来的，那碗药也是夜四小姐亲自喂的，太子殿下……说不定也是知情的。”	}&	She's a present personally sent to us by the Fourth Young Miss of the Ye family. That bowl of medicine was also personally fed to her by the Fourth Young Miss…so this matter…may also have been known by His Highness.	&	"Small waste is the night four Miss personally sent, that bowl of medicine is also the night four Miss personally fed, His Royal Highness ... ... Maybe it's also informed. "	\\
8) & \ZH{	妈哒！ 这是个什么鬼情况！	}&	Damn it, what kind of crappy drama is happening?	&	Damn! What a ghost situation!	\\
9) & \ZH{	\redEnt{夜清落}紧蹙着细眉，努力的睁开沉重的眼皮。	}&	\redEnt{Ye Qing Luo} scrunched \redEntOmit{her} brows together, mustering all her energy to lift her heavy eyelids.	&	\redEnt{The night fell} with a thin brow, and tried to open the heavy eyelids.	\\
10) & \ZH{	刚一睁开，就被极为耀眼的光芒，刺的她眼皮一痛。	}&	Just as \redEntOmit{she} opened them, \redEnt{her} eyes were stung by a bright light.	&	As soon as \redEntOmit{she} opened, she was the bright light, stabbing \redEnt{her} eyelids a pain.	\\
11) & \ZH{	一幕幕陌生的画面，宛如走马灯在脑海里不断的回旋。 	}&	Suddenly, \redEntOmit{her} mind reeled and it felt as if a there was an explosion in \redEntOmit{her} head. Fragments of unfamiliar pictures and scenes started to flood \redEntOmit{her} mind. 	&	A scene of strange scenes, like walking horse lights in the mind of the constant swing. 	\\
12) & \ZH{	斑斓画面一过，那些景象，像是强行插入的记忆，快速的在脑海里重叠，旋即渐渐归于平静。	}&	It continued to flash in \redEntOmit{her} mind non stop when \redEntOmit{she} suddenly realized that these fragments were forcing themselves into \redEntOmit{her} own memories as they melded and fused together. Soon, everything was calm.	&	The scene, those scenes, like forced insertion of memories, quickly overlapped in the mind, and gradually fell calm.	\\
13) & \ZH{	接收完这些记忆后，\redEnt{夜清落}再次睁开了眼睛。	}&	After \redEntOmit{she} received these memories, \redEnt{Ye Qing Luo} tried to pry open \redEntOmit{her} eyes again.	&	After receiving these memories, \redEnt{the night fell} and opened \redEntOmit{his} eyes again.	\\
14) & \ZH{	这一次，\redEnt{她}的眼睛适应了屋内的烛光摇动，灯火明耀。	}&	This time, \redEnt{her} eyes adapted quickly and focused on the candles.	&	This time, \redEnt{her} eyes adapted to the candlelight in the house, and the lights lit up.	\\
15) & \ZH{	奢华精致的房间，颇有古风意味，白色纱帐随风浮动。	}&	It was a luxurious room, exquisitely decorated in an ancient flavour.	&	Luxurious and sophisticated rooms, quite ancient, white yarn book with the wind floating.	\\
16) & \ZH{	四颗夜明珠伫立在房间四个角落，散发着莹莹的光芒。	}&	Four night pearls stood on each corner of the room, illuminating the room aglow.	&	Four night pearls stand in the four corners of the room, emitting a bright light.	\\
17) & \ZH{	最忒玛坑爹的是，\redEnt{她}现在四肢大敞的躺在一张圆木桌上，自己原本那具引以为傲的身材，变成了纤细娇弱的少女身体，只着了一件白色长袍。	}&	What seemed even more shocking was that \redEnt{she} was now lying atop a wooden table, \redEntOmit{her} voluptuous figure was now instead a slender and delicate girl's body, with only a white robe on.	&	The most beautiful father is that \redEnt{she} is now lying on a round wooden table with \redEntOmit{her} limbs, \redEntOmit{her} original proud figure, into a delicate girl's body, only a white robe.	\\
18) & \ZH{	靠！ \redEnt{她}这样躺在桌子上，是当做食物，等待着别人把\redEnt{她}吃掉吃掉吃掉咩？	}&	What the?! Lying on the table like this seemed to be as if \redEnt{she} was some kind of food and waiting for someone to eat \redEnt{her} up?	&	By! \redEnt{She} lay on the table like this, as food, waiting for others to eat \redEnt{her} and eat \redEnt{her}?	\\
19) & \ZH{	尼玛\redEnt{她}不就是从飞机上跳了下来，就把自己的灵魂转移到了这么一具小屁孩的身体里了？	}&	Damn it! \Verb{Didn’t} \redEnt{she} jump off the plane? How did her soul end up in this little girl’s body?	&	\Verb{Does} Neymar just jump off the plane and transfer her soul to such a little?	\\
20) & \ZH{	而且，刚穿越就这么惨，还能不能愉快的玩耍了？	}&	And since \redEntOmit{she}’s been given a new lease of life, \Verb{wasn’t} this start a bit too miserable?	&	And, just through such a miserable, \Verb{can} not be happy to play?	\\
21) & \ZH{	\redEnt{夜清落}想动，使不上劲。	}&	\redEnt{Ye Qing Luo} tried to move again, but \redEntOmit{she} could not muster any strength at all.	&	\redEnt{The night fell} and tried to move, so that \redEntOmit{it} could not be vigorous.	\\
22) & \ZH{	门外的那两人刚刚说过，\redEnt{她}被人下了药。	}&	As what the two men had said earlier, \redEnt{she} was drugged.	&	The two men outside the door had just said that \redEnt{she} had been drugged.	\\
23) & \ZH{	而且下药的那人，是这具身体的四姐，并且亲自把\redEnt{她}送到了这里，让人来玷污\redEnt{她}！	}&	And what’s more, the drug was personally administered by \redEntOmit{her} very own sister and even sent \redEnt{her} here as a present to these men to tarnish \redEnt{her}!	&	And the man who took the medicine, is the body of the four sisters, and personally sent \redEnt{her} here, let people to tarnish \redEnt{her}!	\\
24) & \ZH{	\redEnt{夜清落}快速从记忆中搜寻自己所需要的记忆。	}&	\redEnt{She} quickly searched through \redEntOmit{her} memories.	&	\redEnt{Night clearing} quickly searches for the memories \redEntOmit{you} need from memory.	\\
25) & \ZH{	门外那人，是玄者四大家之一\bleuEnt{尉迟}世家的四少爷尉迟涯，\bleuEnt{此人}风流成性，游手好闲，就是一个大写的纨绔少爷。	}&	The man who was outside, was from the \bleuEnt{Yuchi} clan, one of the four major family clans. \bleuEnt{He} was the Fourth Young Master of the \bleuEnt{Yuchi} family, a well known foppish playboy who spent, \bleuEntOmit{his} time idling about - \bleuEnt{Yuchi Ya}.	&	The person outside the door, is one of the four people of the \bleuEnt{Xuan}, the captain of the late family of the four young captain siaa, this person is a popular, idle, is a capital is a master.	\\
26) & \ZH{	把\redEnt{她}送到\bleuEnt{尉迟涯}面前，根本就是送羊入“狼”口！	}&	Sending her to \bleuEnt{Yuchi Ya} was simply putting a sheep in front of a wolf’s mouth!	&	Send \redEnt{her} to the captain in front of the \bleuEnt{late ya}, is simply to send sheep into the "wolf" mouth!	\\
27) & \ZH{	四姐？ 还有那个所谓的未婚夫？	}&	Fourth Sister? And who was her so called fiance?	&	Four sisters? And that so-called fiance?	\\
28) & \ZH{	呵！	}&	Ah!	&	Oh!	\\
29) & \ZH{	都给她等着！	}&	They better be good and wait for her to return this favour back many folds!	&	Just give her a wait!	\\
30) & \ZH{	\redEnt{夜清落}微眯起锋锐的眼睛，强压住身体传来的剧痛和麻木，努力的控制着四肢。	}&	\redEnt{Ye Qing Luo}'s gaze sharpened and she exerted a strong pressure, using all her effort to regain control of her limbs.	&	\redEnt{The night fell} slightly with sharp eyes, pressed the body from the sharp pain and numbness, and tried to control the limbs.	\\
31) & \ZH{	“吱呀”一声门响，\bleuEnt{尉迟涯}走了进来。	}&	[Squeak-] The door opened and \bleuEnt{Yuchi Ya} strode in.	&	"Squeaky" a door rang, \bleuEnt{the captain} came in late.	\\
32) & \ZH{	听脚步声，少说也有五人以上。	}&	From the sound of the footsteps, \redEntOmit{she} gathered that there were at least five or more people with \bleuEnt{him}.	&	Listen to the footsteps, less say there are more than five people.	\\
33) & \ZH{	“\redEnt{小废物}，\bleuEnt{哥哥}现在就来疼你！”	}&	\redEnt{Little Waste}, \bleuEnt{brother} is here to dote on you… 	&	"Little waste, \bleuEnt{brother} is here to hurt you now!" 	\\
34) & \ZH{	\bleuEnt{尉迟涯}走到桌子边，直接伸手扯\redEnt{她}身上的白袍。	}&	\bleuEnt{He} leered and slowly walked over to the table and immediately reached for \redEnt{her} white robe.	&	\bleuEnt{The captain} walked up to the table and reached directly for \redEnt{her} white robe.	\\
35) & \ZH{	\redEnt{夜清落}冰冷的眼神锐利，沙哑着嗓音，吐出一个字：“滚！” \bleuEnt{尉迟涯}听到\redEnt{她}的声音，笑得更是嚣张：“还没昏死过去？也好，也好！”	}&	When \bleuEnt{Yuchi Ya} heard \redEnt{her} voice, \bleuEntOmit{he} laughed even more lasciviously, with a hint of arrogance, "You’re awake? Very good, very good!"	&	\redEnt{The night clear} cold eyes sharp, hoarse voice, spit out a word: "Roll! When \bleuEntOmit{he} heard her voice, \bleuEntOmit{he} smiled more loudly: "Haven't passed out yet? Good, good!"	\\
\bottomrule[2pt]
\end{tabular}
\end{adjustbox}
\caption{Another example chapter in \BWB. This example is even more difficult for \MT{} since it fails to recognise the main character ``Ye Qing Luo'' as a named entity.}
\label{fig:example_book153_0}
\vspace{-5pt}
\end{figure*}

\end{appendices}

\end{document}